\documentclass[10pt,twocolumn,letterpaper]{article}

\usepackage{iccv}
\usepackage{times}
\usepackage{epsfig}
\usepackage{multirow}
\usepackage{adjustbox}
\usepackage{subfig}
\usepackage{graphicx}
\usepackage{amsmath}
\usepackage{amssymb}
\usepackage{pdflscape}

\newcommand*\samethanks[1][\value{footnote}]{\footnotemark[#1]}

\usepackage[breaklinks=true,bookmarks=false]{hyperref}

\iccvfinalcopy 

\def\iccvPaperID{****} 

\ificcvfinal\fi

\begin{document}

\title{CIGLI: Conditional Image Generation from Language \& Image}

\author{
Xiaopeng Lu\thanks{ Equal Contribution} \qquad Lynnette Ng\samethanks \qquad Jared Fernandez\samethanks \qquad Hao Zhu\samethanks\\
Carnegie Mellon University\\
{\tt xiaopen2@cs.cmu.edu, \{lynnetteng, jaredfern, zhuhao\}@cmu.edu }
} 


\maketitle
\ificcvfinal\thispagestyle{empty}\fi


\begin{abstract}
    Multi-modal generation has been widely explored in recent years. Current research directions involve generating text based on an image or vice versa. In this paper, we propose a new task called CIGLI: Conditional Image Generation from Language and Image. Instead of generating an image based on text as in text-image generation, this task requires the generation of an image from a textual description and an image prompt. We designed a new dataset to ensure that the text description describes information from both images, and that solely analyzing the description is insufficient to generate an image. We then propose a novel language-image fusion model which improves the performance over two established baseline methods, as evaluated by quantitative (automatic) and qualitative (human) evaluations. The code and dataset is available at \url{https://github.com/vincentlux/CIGLI}.
\end{abstract}

\section{Introduction}
Multimodal reasoning over visual and language is a long-standing research problem with the ultimate goal of building a system to connect and understand information across two vastly different inputs. Current research in this area primarily focuses on classification tasks, where the model jointly understands the text and image information to produce a classification label. State-of-the-art multimodal models separately model the caption and the images before jointly processing the combined representation to perform a \textit{TRUE/FALSE} evaluation ~\cite{tan2019lxmert,su2019vl, chen2020uniter}.

One such discriminative multimodal reasoning task is Natural Language Visual Reasoning (NLVR) \cite{suhr2017corpus} which defines a binary classification task to test model's reasoning capabilities across modalities. Given a sentence description and three synthetic images, the task is to predict whether the sentence correctly described the images or not. In NLVR2 \cite{suhr2018corpus}, realistic natural images are added though the task remains the same. 

While advances in modeling have led to improved results on these tasks, current models are largely based around BERT-style autoencoding architectures which are not well-suited for generation. As opposed to classification tasks which only require models to output a label, image generation from natural language captions presents further challenges in the multimodal regime as it requires a much richer output than a single discriminative label \cite{chen2020generative,pmlr-v37-gregor15,reed2016learning}. 
In this work, we propose a novel generative task: given a description describing two images, and one of the images, can a model learn to generate the second image that is \textit{semantically} correct? For example, given a text description ``There are two dogs in total", and a first image with one dog in it, the model should be able to generate the second image with exactly one dog. While recent work has explored conditional generation of images from natural language \cite{ramesh2021zero}, to our knowledge there has been no prior work constraining generation based on both language and visual inputs. Our task provides the model with the caption and the first image in order to generate the second image, requiring the examination of both the language and visual prompts and understand the semantic relationships between the input text and image. This task requires the model to both understand the natural language caption and identify the missing visual components from the first image to generate the second image in the pair. Thus, it cannot be performed by conditioning solely on text as the contents of the second image are directly dependent on the contents of the first.


We construct a new dataset based off the NLVR2 task \cite{suhr2018corpus} and perform image generation on two GAN-based architectures. We further evaluate their performance using both quantitative (automatic) and qualitative (human) evaluations. While our results are preliminary, we hope to inspire further directions in language-visual reasoning, especially in the combined understanding of both mediums.

\section{Related Work}
Many neural approaches to text-to-image generation rely on architectures built around Generative Adversarial Network (GANs). Several GAN-based models composed images by iteratively retrieving word vectors for different portions of the image and  up-scaling the image resolution \cite{Tao18attngan,han2017stackgan}. Alternate GAN-based approaches attempted different combinations of inputs to the generators and discriminators, such as: text features inputs into both generator and discriminator \cite{pmlr-v48-reed16}, text and image features input both the generator and discriminator \cite{Yizhe_ZSL_2018}, or text features into the generator and image into the discriminator \cite{tao2021dfgan}. 
Recent text-to-image models have also been developed around the transformer architectures such as  X-LXMERT~\cite{cho2020xlxmert} and DALL-E \cite{ramesh2021zero}.


\section{Dataset Construction}
We construct a new dataset by combining the train and dev splits of the original NLVR2 dataset and filtering for image pairs and captions which require visual reasoning. Each data point in the source NVLR2 dataset contains two images and a natural language sentence. Next, to ensure that the generation task requires reasoning about both the language and visual prompt, we removed all \textit{FALSE} examples, as negative captions can trivially be satisfied by outputting noise for the generated image.

\begin{figure}[!t]
    \centering
    \includegraphics[width=0.4\textwidth]{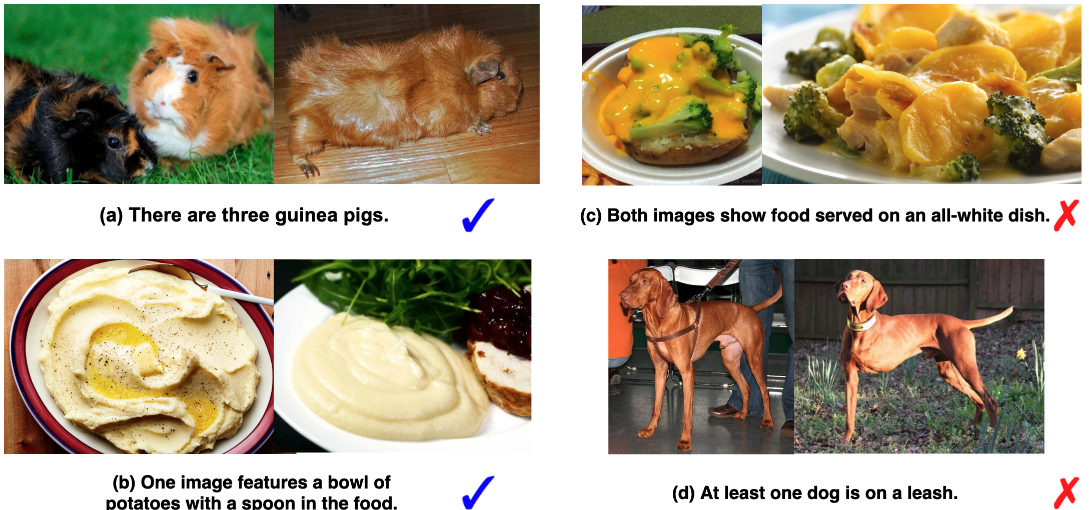}
    \caption{Cases of qualified and unqualified data. (a) is qualified because counting about the number of objects across both images. (b) is qualified because it requires distinguishing which textual condition is satisfied by the first image before generating the second image. (c) is unqualified because the second image can be generated without looking at the first. (d) is unqualified because the model can ignore the first image and generate a single dog.}
    \label{fig:qual-example}
\end{figure}

Additionally, we manually examined 600 randomly sampled images to determine a set of heuristic patterns found in the text captions that require reasoning about both images. These patterns consist of captions where the model cannot perform image generation based on the caption alone; it must also analyze the first image to determine which part of the caption it satisfies. Figure~\ref{fig:qual-example} shows examples of filtered data. 
The final dataset consists of 24,178 valid data points that were found to satisfy our heuristic-based filtering \footnote{filtering code is available at \url{https://github.com/vincentlux/CIGLI/blob/main/filter.py}}. 


\section{Methodology}
\begin{figure}[!ht]
    \centering
    \includegraphics[width=0.5\textwidth]{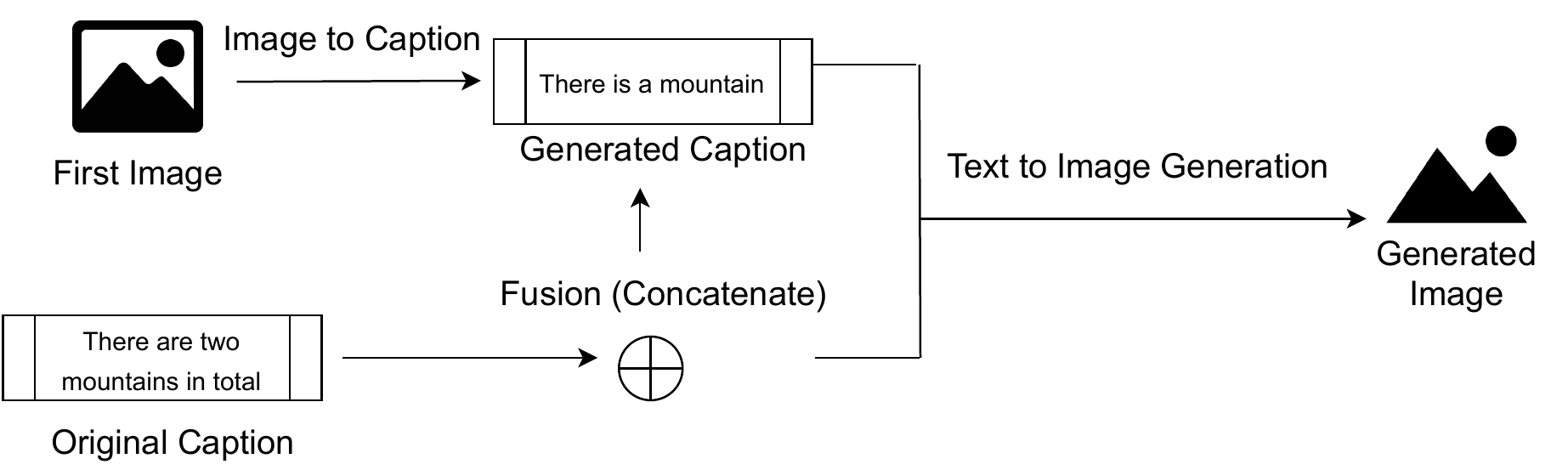}
    \includegraphics[width=0.5\textwidth]{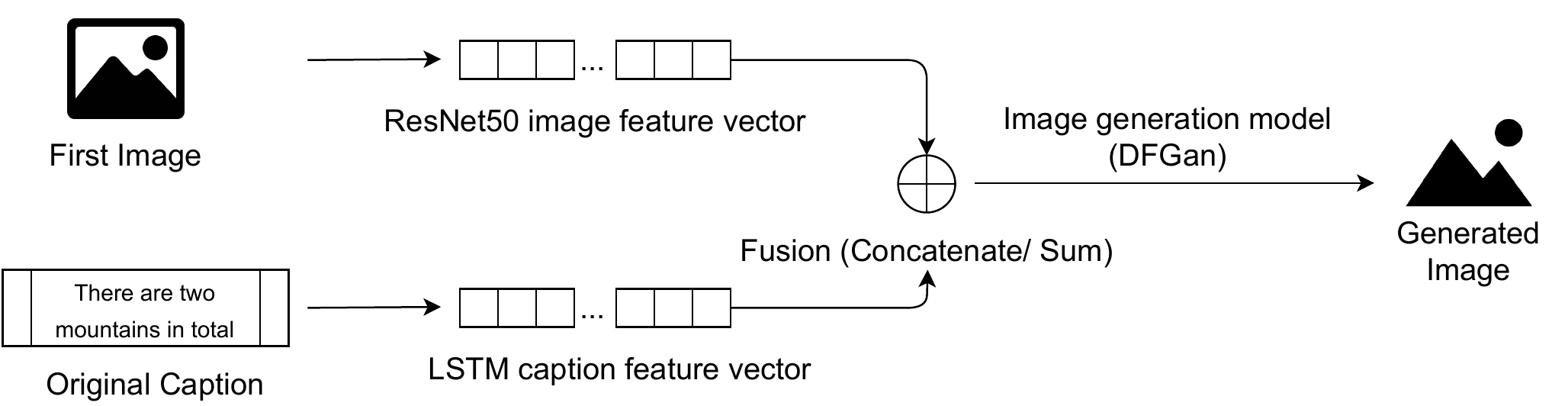}
    \caption{Baseline model (upper image) vs. Our model (lower image)}
    \label{fig:baselinemodel}
\end{figure}

In this section, we outline our methodology for image generation baselines task with our constructed dataset. We first perform an evaluation using language-only baseline models from available pre-trained AttnGAN and DF-GAN models, then select the better of the two for further improvements. We improve the model through fine-tuning the model and fusion of image features. The two model architectures are reflected in Figure ~\ref{fig:baselinemodel}.

\subsection{Baseline Method}
We use a language-only model for our baseline image generation. The model is fed an initial image, named the ``First Image", and the original caption that describes both images. From the first image, we generate a caption for it through an image to caption algorithm. We then concatenate the original caption with the generated caption, and pass it through a text-to-image-generation algorithm to produce the final generated image.

In the first step, image captioning \cite{kiros2014multimodal} translates images into text in two steps: (1) image encoding; (2) language generation. We use the following image captioning model: LSTM \cite{hochreiter1997long} with features from pretrained ResNet50 \cite{he2016deep}, which achieves 35.9 BLEU@4 on MSCOCO \cite{lin2014microsoft}. We use beam search (beamSize=10) to get the best captions for each image. We generate 10 captions for each image through this algorithm. 

We then concatenate all ten of the generated captions together with the original NLVR caption in order to get as much information for the text-to-image generation step. This becomes the input for the image generation model. We run image generation on two pre-trained models: AttnGAN \cite{Tao18attngan} and DF-GAN \cite{tao2021dfgan}. Both models were worked on based on a pre-trained MSCOCO model. Although these models have better output with other datasets or subsets of MSCOCO, we chose to use the entire dataset as it has more diverse objects, in line with our natural image generation task. AttnGAN was run with a text embedding vector of size 256, and a training batch size of 100, and produces an image of a 256 * 256 resolution. 


\subsection{Proposed Method}
We perform automatic evaluations on baseline generated images and improved on DF-GAN, as it is the better of the two models. We use a language-and-vision model for generation (Figure \ref{fig:baselinemodel}, lower image). We represent the First Image as a vector from ResNet50 features pretrained on MSCOCO. We represent the text caption as an LSTM vector. We then concatenate both image and language vectors, and input the combined representation into an image generation model. We use DF-GAN as our image generation model to provide direct comparison with our baseline model. We employ two fusion techniques: (1) concatenation of the image and text feature vectors and (2) sum of the image and text feature vectors.

We train our proposed network end-to-end without freezing the text or image encoder sub-networks, with 120 epochs and a batch size of 24.


\subsection{Evaluations}
We employ automatic evaluation using the UNITER metric and the Inception Score metric, and human evaluation. We used automatic evaluation to evaluate our baseline models of AttnGAN and DF-GAN, before picking the better model to improve on; then used both automatic and human evaluation to evaluate our proposed model architecture.

\subsubsection{Automatic Evaluation} 


\textbf{UNITER metric}
To provide an automated evaluation of the generated image quality, we sought to determine if the generated images were sufficient for a fine-tuned UNITER model \cite{chen2020uniter} to recognize that the pair of images satisfied the reasoning statement in the caption.

We performed visual feature extraction using a Faster RCNN network to extract 768-dimensional visual features from both the generated images and the source images. Prior to use as a metric, the UNITER model was finetuned on the NLVR2 task for 8000 steps. The UNITER model was then provided 1805 true examples in which the caption correctly describes the image pairs. To compare against existing generative baselines, we swapped out the second image in the pair with the generation in the data point.  

\textbf{Inception Score}
To understand the quantitative quality of the image, we use the Inception Score \cite{salimans2016improved}. Although the quality of image speaking of naturalness and diversity is not our main concern, inception score serves to evaluate the general quality of the generated image. 

\subsubsection{Human Evaluation}
Quantitative intrinsic metrics fail to capture the semantic correctness of the generated image. Therefore, human evaluation plays a significant role in our experiment. Based on previous research~\cite{ramesh2021zero}, we define three dimensions for human evaluation:

\textbf{Naturalness}: the evaluators were asked to use their intuition on how natural the generated image looks

\textbf{Relevance} whether the images are relevant to the caption

\textbf{Correctness}: whether the generated image is semantically correct (eg. number of objects, object type etc)


We constructed a web application for human evaluation. For each data point, the evaluators are presented with the caption and the First Image. They are asked if the image satisfies the caption. If they select ``NO", they are presented with the generated images to evaluate further. Evaluators are presented with the generated images one by one and asked to provide a YES/NO binary rating to each of the metric. This is a blind evaluation; the evaluators do not know which model generated which image. We did not show the model name to the evaluators to alleviate the bias related to knowing how the image is generated. Figure in the appendix shows an example of a data point on the application presented to the evaluators for each type of rating scheme.

\section{Results}
\begin{figure*}[!ht]
\small
\begin{center}
  \begin{adjustbox}{max width=0.9\textwidth}
\begin{tabular}{|p{3cm}|p{2.5cm}|p{2.5cm}|p{2.5cm}|p{2.5cm}|p{2.5cm}|p{2.5cm}|}
\hline
\textbf{Caption} & \textbf{First Image} & \textbf{NLVR2 Second Image} & \textbf{AttnGAN baseline} & \textbf{DF-GAN baseline} & \textbf{DF-GAN Concat} & \textbf{DF-GAN sum} \\ \hline 
One image features puppies next to an adult dog& \includegraphics[width=0.8\linewidth]{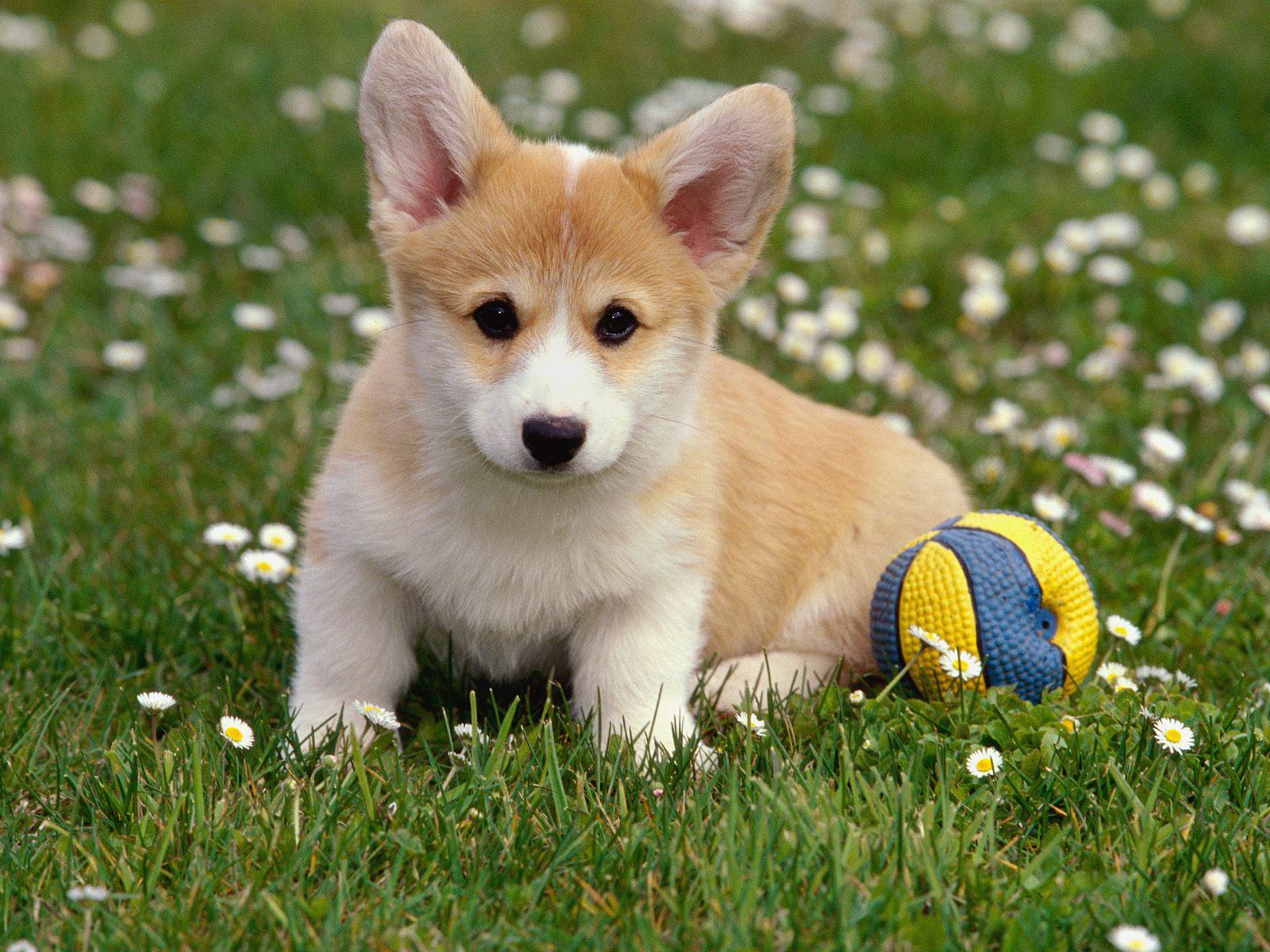} & \includegraphics[width=0.8\linewidth,height=0.8\linewidth]{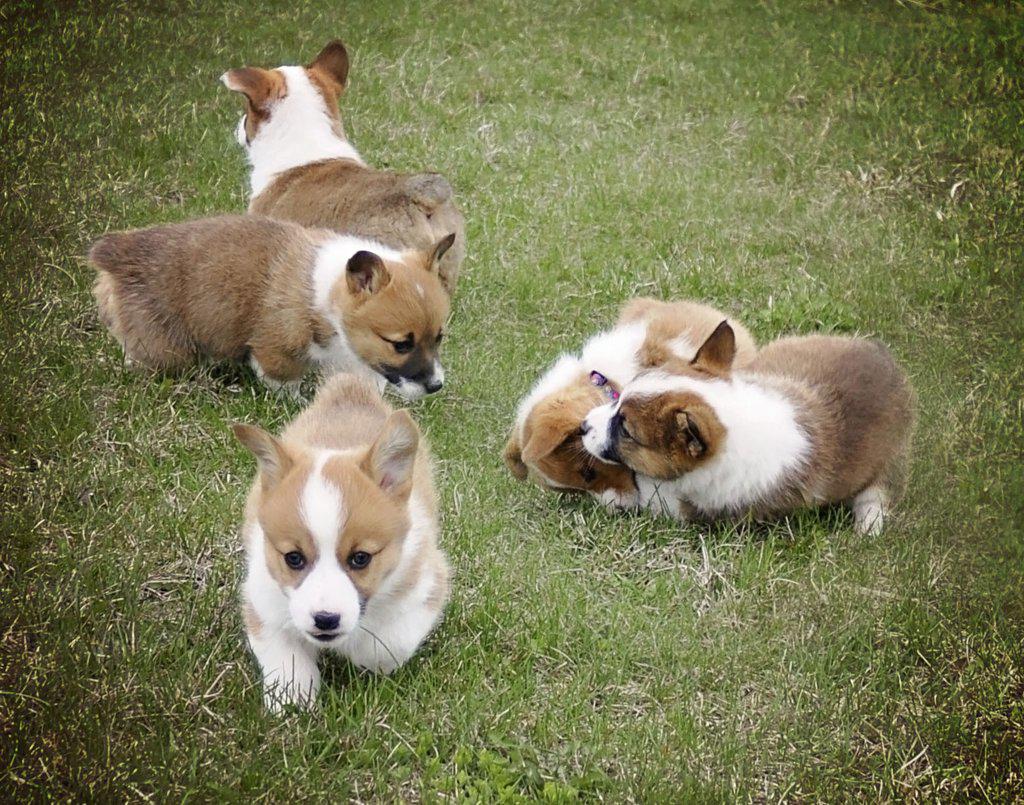} & \includegraphics[width=0.8\linewidth]{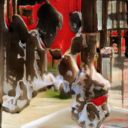} & \includegraphics[width=0.8\linewidth]{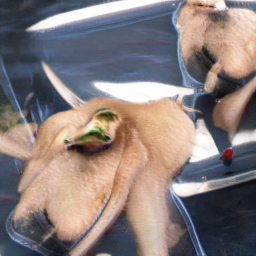} & \includegraphics[width=0.8\linewidth]{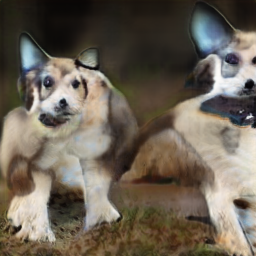}& \includegraphics[width=0.8\linewidth]{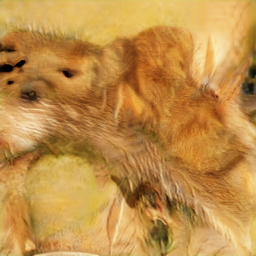} \\ \hline
There are exactly two dogs & \includegraphics[width=0.8\linewidth]{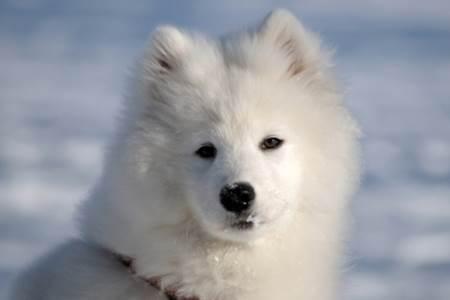}  & \includegraphics[width=0.8\linewidth,height=0.8\linewidth]{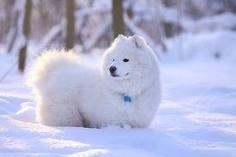} & \includegraphics[width=0.8\linewidth]{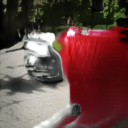}  &\includegraphics[width=0.8\linewidth]{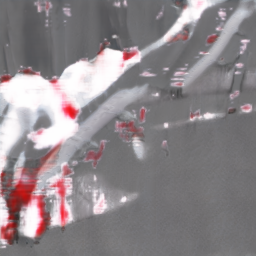} & \includegraphics[width=0.8\linewidth]{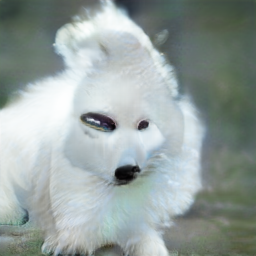}& \includegraphics[width=0.8\linewidth]{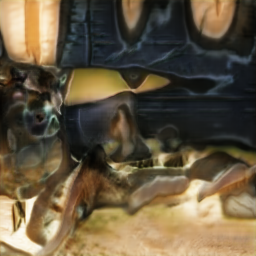} \\ \hline
The right image contains exactly one gorilla & \includegraphics[width=0.8\linewidth]{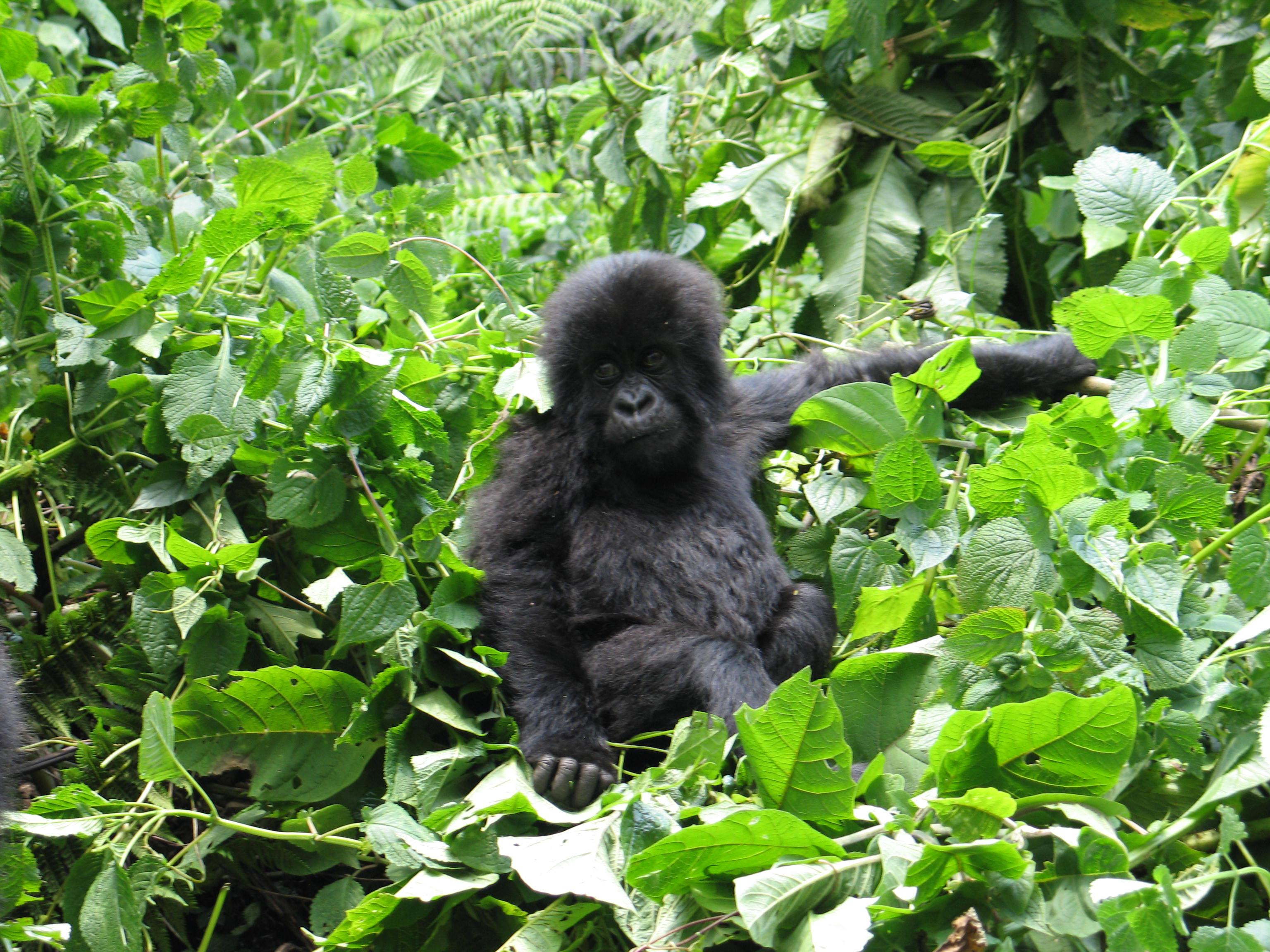}  & \includegraphics[width=0.8\linewidth,height=0.8\linewidth]{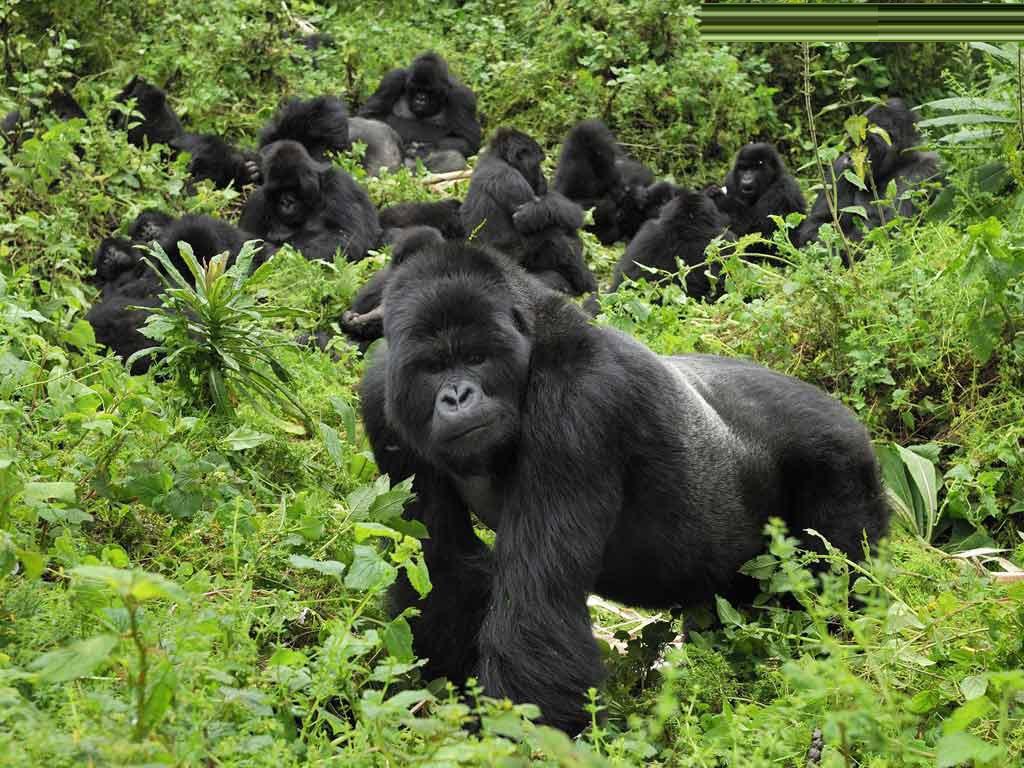} & \includegraphics[width=0.8\linewidth,height=0.8\linewidth]{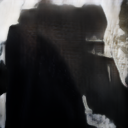}  &\includegraphics[width=0.8\linewidth]{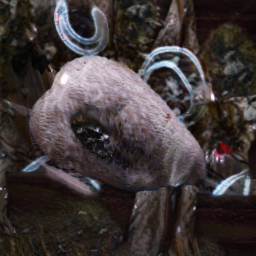} & \includegraphics[width=0.8\linewidth]{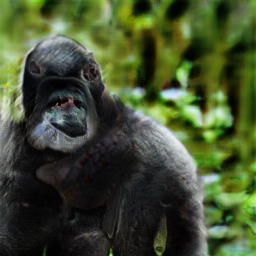}& \includegraphics[width=0.8\linewidth]{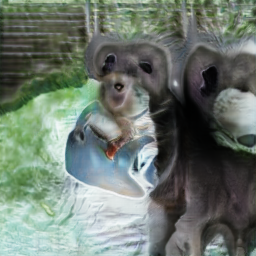} \\ \hline
\end{tabular}
\end{adjustbox}
\caption{Qualitative results}
\label{tab:qualitativeresults}
\end{center}
\end{figure*}

The generated images are presented in Table ~\ref{tab:qualitativeresults}. The automated evaluations using UNITER and Inception Score results are in Table ~\ref{tab:inception+uniter}. 

\begin{table*}[]
    \centering
    \small
  \begin{adjustbox}{max width=0.8\textwidth}
    \begin{tabular}{l|c|c|c}
        & Inception Score$\uparrow$ & UNITER-Accuracy$\uparrow$ & UNITER-Consistency$\uparrow$  \\  \hline
        Original Image & - & 76.07 & 64.96\\ \hline
        \multicolumn{3}{l}{Baseline Models} \\ \hline
        AttnGAN & 5.50 $\pm$ 0.43 & 41.55 & 25.13\\  
        DF-GAN & 9.59 $\pm$ 0.59 & 42.94 & 26.25 \\ \hline
        \multicolumn{3}{l}{Proposed Models} \\ \hline
        DF-GAN Concatenated & \textbf{11.30} $\pm$ 0.83 & \textbf{46.54} & 28.50 \\ 
        DF-GAN Sum & 9.67$\pm$0.53 & 44.97 & \textbf{28.80} \\ \hline
    \end{tabular}
  \end{adjustbox}
    \caption{Automatic Evaluation Score Comparison}
    \label{tab:inception+uniter}
\end{table*}

\begin{figure}[!h]
    \centering
    \includegraphics[width=0.4\textwidth]{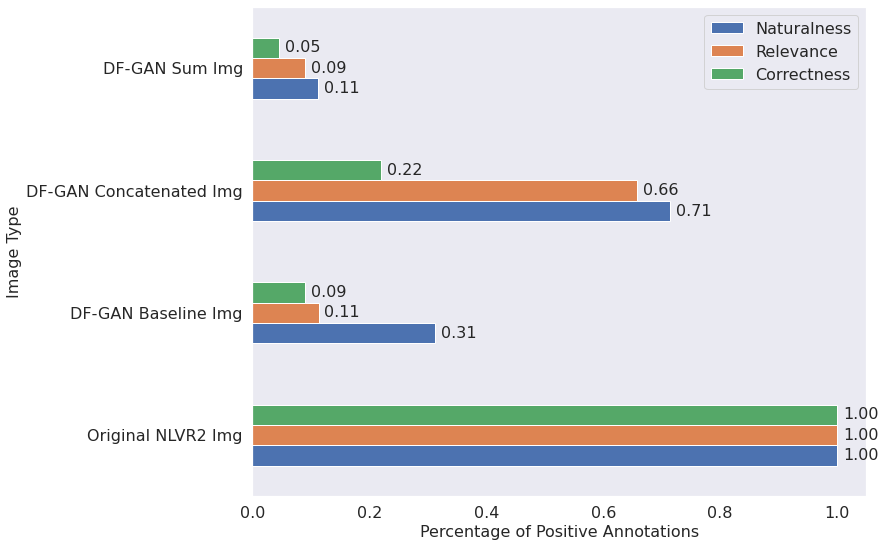}
    \caption{Human evaluation results}
    \label{fig:humanevalfinalscheme}
\end{figure}

For human evaluation, we used four evaluators. The results are presented in Figure \ref{fig:humanevalfinalscheme}. The evaluators identified that in 34\% of the data points presented, the first image had already satisfied the caption. We then evaluated the remaining 66\% of images for their naturalness, relevance and accuracy. We evaluated the percentage of positive annotations for each metric; i.e. the percentage where evaluators annotate YES to the metric. As expected, the NLVR2 image ranks at 100\% for all three categories. The concatenated fusion model (DF-GAN concatenated) produces higher human ratings as compared to the summation fusion model (DF-GAN sum). These results are consistent with the automated evaluation, where DF-GAN Concatenated image reports the best result quality. 

\section{Discussion}
Previous work in image generation typically generate an output image from a single text caption. In this study, our image generation model is conditioned on both an input image to and a text caption. Our proposed DF-GAN concatenated fusion model outperforms the sum-fusion and the baseline models, suggesting that the images generated from this architecture are more similar to the original image than other architectures. Still, the image generation output has a long way to go if we compare the score with original NLVR2 original images, showing that CIGLI task is far from been solved, and more future work can be done on this task to improve the performance.

\paragraph{Automatic evaluation score analysis}
The accuracy of the UNITER model on the ground truth original image serves as an upper bound baseline for the performance of generated images. The DF-GAN network trained with our proposed technique outperforms both language-only AttnGAN and DF-GAN baselines on both accuracy and consistency. This suggests that the images generated with our proposed metric are better able to reflect the necessary visual features needed for the UNITER model to recognize that the image pair is a valid reasoning statement. 

Inception score reflects the naturalness and diversity of the generated images. Although this metric cannot measure the semantic correctness of the image, it can still be used as a way to help understand how good is the generation quality in general. From our results, our proposed concatenated model improves the Inception score by 1.71 compared with the DF-GAN baseline. We posit this is due to the extra NLVR2 training data we used for conditional image generation.

\paragraph{Human evaluation score analysis}
The human evaluation results shows promise as our proposed methods outperforms the baseline method. Despite the lack of naturalness and correctness of our proposed models as compared to the original NLVR2 images, a manual inspection observes that our method performs pretty well on common objects such as dogs, and not so well on uncommon objects like alpacas. Captions that tend to go along with good output images are quantitative captions (eg ``there are two dogs") describing a single object, rather than a comparison of images (eg ``the right image contains... while the left image contains...") or captions describing multiple objects in an image (eg ``there is a girl using a smartphone").


\section{Conclusions}
In this paper, we introduce CIGLI, a new task for conditional image generation from both text and images. This task poses new challenges to vision-language research, as it requires model to perform semantic reasoning across both modalities before image generation. We also propose a new image-text fusion model based on DF-GAN, which improves the performance compared with two baseline models. We hope that this new dataset and the corresponding model can be the start point of future research in conditional image generation from language and image.

{\small
\bibliographystyle{ieee_fullname}
\bibliography{
    egbib,
    acl2020, 
    anthology, 
    related_work,
    baselines
}

}

\end{document}